\documentclass[conference]{IEEEtran}
\IEEEoverridecommandlockouts
\usepackage{cite}
\usepackage{amsmath,amssymb,amsfonts}
\usepackage{algorithmic}
\usepackage{graphicx}
\usepackage{textcomp}
\usepackage{xcolor}
\usepackage[T1]{fontenc}
\usepackage[utf8]{inputenc}
\usepackage{graphicx}
\usepackage{CJKutf8}
\usepackage{bbding}
\usepackage{multirow}
\usepackage{booktabs}
\usepackage{caption}
\usepackage{array}
\usepackage{arydshln}
\definecolor{blueaccent}{RGB}{240,221,157}
\definecolor{greenaccent}{RGB}{149,200,180}
\definecolor{purpleaccent}{RGB}{130,41,128}
\definecolor{orangeaccent}{RGB}{240,83,50}
\usepackage{hyperref}
\hypersetup{
    colorlinks=true,
    linkcolor=blue,
    filecolor=magenta,      
    urlcolor=cyan,
    pdftitle={Overleaf Example},
    pdfpagemode=FullScreen,
    }
\usepackage{cleveref}
\usepackage{graphicx}    
\usepackage{tikz}
\usepackage{pgfplots}
\usepackage{bchart}
\usetikzlibrary{decorations.pathreplacing}
\usepackage{subfig}
\pgfplotsset{compat=1.10}
\usepackage{pgfplotstable}
\usepackage{microtype}

\usepackage{inconsolata}

\newcommand{\llms}{{LLMs} }

\def\BibTeX{{\rm B\kern-.05em{\sc i\kern-.025em b}\kern-.08em
    T\kern-.1667em\lower.7ex\hbox{E}\kern-.125emX}}

\begin{document}

\title{FFN: a Fine-grained Chinese-English Financial Domain Parallel Corpus}


\author{\IEEEauthorblockN{Yuxin Fu,  ~~~~ Shijing Si$^*$\thanks{\IEEEauthorrefmark{1}Corresponding author: Shijing Si, \texttt{shijing.si@outlook.com}.},}
\IEEEauthorblockA{\textit{School of Economics and Finance} \\
\textit{Shanghai International Studies University}\\
Shanghai, China}
\and
\IEEEauthorblockN{Leyi Mai, ~~~~ Xi-ang Li}
\IEEEauthorblockA{\textit{School of Education} \\
\textit{Shanghai International Studies University}\\
Shanghai, China}
}

\maketitle

\begin{abstract}
Large Language Models (LLMs) have stunningly advanced the field of machine translation, though their effectiveness within the financial domain remains largely underexplored. To probe this issue, we constructed a fine-grained Chinese-English parallel corpus of financial news called FFN. We acquired financial news articles spanning between January 1st, 2014, to December 31, 2023, from mainstream media websites such as CNN, FOX, and China Daily. The dataset consists of 1,013 main text and 809 titles, all of which have been manually corrected. We measured the translation quality of two LLMs -- ChatGPT and ERNIE-bot, utilizing BLEU, TER and chrF scores as the evaluation metrics. For comparison, we also trained an OpenNMT model based on our dataset. We detail problems of LLMs and provide in-depth analysis, intending to stimulate further research and solutions in this largely uncharted territory. Our research underlines the need to optimize LLMs within the specific field of financial translation to ensure accuracy and quality.
\end{abstract}


\section{Introduction}



Translation in business and finance domain has increased in volume as well as impact due to the growing globalisation and explosion of financial transactions and increasing business activity \cite{biel2017translation}. 
China are the first populous nations of the world, and arguably, the growth in wealth and spending power of it make it very attractive destinations for business \cite{Turenne2022mining}.  Besides, English is the dominant language in the global business \cite{hung2002translation}. Therefore, the demand for Chinese-English translation is huge, across many areas and industry sectors. Due to
the complicated nature of financial concepts, translators have to commit to a big up-front investment in order to acquire a deep knowledge of the various sub-sectors and types of texts, each with a different level of lexical complexity. 

Large language models (LLMs) pretrained on massive unlabeled corpora have shown impressive emergent abilities under model scaling which enable prompting for downstream applications~\cite{gpt3,kaplan2020scaling,wei2022emergent,zhang2022opt,chowdhery2022palm}.
However, there is little work on exploring the application of \llms for machine translation in the financial domain.

Additionally, the translation performance of \llms is derived from their training datasets. If we want to study the effectiveness of large language models in Chinese-English translation within the financial domain, it is also essential for us to search for existing datasets in this field.
Based on the above considerations, we conducted relevant research and experiments. In this paper, our main contributions are listed as below:
\begin{itemize}
    \item We build a parallel dataset of English-Chinese news translation in the finance domain, which includes main texts and titles.
    \item Based on our parallel dataset, we evaluated the performance of ChatGPT and ERNIE-bot in translation, and brought in DeepL and Google for comparison, and found some unexpected feedback.
    \item We trained an OpenNMT model based on it to evaluate the performance of the dataset.
    \item We also provide a quantitative and qualitative analysis to disclose problems when prompting for MT, which provides insights for future study.
\end{itemize}

\section{Related Works}

\subsection{Large Language Models}

Large language models have good promise for machine translation. Reference \cite{moslem2023adaptive} found that \llms have recently shown interesting capabilities of in-context learning and can adapt to a set of in-domain sentence pairs and/or terminology while translating a new sentence. Reference \cite{xu2023paradigm} proposed a novel fine-tuning approach for LLMs that is specifically designed for the translation task, eliminating the need for the abundant parallel data that traditional translation models usually depend on. When it comes to ChatGPT, reference \cite{hendy2023good} presented a comprehensive evaluation of GPT models for machine translation and found that GPT models achieve very competitive translation quality for high resource languages, while having limited capabilities for low-resourced languages. Although Chinese is one of the high resource languages, the study of Chinese-English translation quality of LLMs is still under-explored. 
In this paper, we release a high-quality, human verified parallel dataset that can benchmark popular LLMs.

\subsection{Datasets}
\begin{table}
\centering
\caption{Detail information of our dataset.
("AVER"is average length,"PARA"is paragraphs)}
\captionsetup{justification=centering}
\scalebox{1.2}{
\begin{tabular}{c|c|c|c|c}
\hline
Language&Category&AVER.&PARA.&TOKEN\\

\hline
Chinese& main texts&123.44&1013&65454\\
&titles&15.70&809&6382\\
\hline
English&main texts&69.41&1013&72509\\
 &titles&8.56&809&7181\\
\hline
\end{tabular}
}

\label{tab:sentence}
\end{table}

\begin{table*}[ht]
\centering

\caption{Three examples in dataset "Financial News dataset for text mining" \cite{Turenne2022mining}.}
\captionsetup{justification=centering}

\begin{tabular}{m{7cm}<{\centering}m{5cm}<{\centering}m{5cm}<{\centering}}
\hline
English Sentences&Chinese Sentences&Notes\\
\hline
 Last month, during London Fashion Week, I took my six-year-old daughter with me to Gareth Pugh's show as a treat. Pugh is one of the more ballyhooed YBDs (Young British Designers), having graduated only three years ago from Central St Martins, received a Topshop-sponsored New Generation grant from the British Fashion Council and garnered lots of attention last season for making clothes that looked like the scene in <em><i>The Incredibles</em> where Mr Incredible tries to escape the bad guy's clutches and gets caught in a trap of inflatable plastic balls.&<div><b>6</b><b>\begin{CJK*}{UTF8}{gkai}岁女孩眼中的\end{CJK*}\&ldquo \begin{CJK*}{UTF8}{gkai}时装\end{CJK*}\&rdquo </b><b></b>&The Chinese and English text clearly don't align or match. Additionally, there are many HTML tags in the text that need preprocessing.\\
\hline
China is leaking and it is probably America’s fault, writes David Keohane. @ China is, after all, the most exposed to the quantity effect of liquidity withdrawal due to the Federal Reserve’s tapering of its quantitative easing programme. @ Indeed, it may be the country that has the most difficulty dealing with it.&\begin{CJK*}{UTF8}{gkai}归根结底，随着美联储(Fed)逐渐退出量化宽松(QE)计划，中国是受流动性撤出的数量效应影响最大的国家。\end{CJK*}&They use "@" to separate sentences, but there are three sentences in English and only one in Chinese. This Chinese sentence corresponds to the second English sentence, while the other two English sentences do not have a corresponding Chinese counterpart. \\
\hline
(lifestyle) Highway to heaven (1044)-28363-28363-28363
&\begin{CJK*}{UTF8}{gkai}大马：西方人的探险乐园
\end{CJK*}&They intended to provide Chinese and English titles corresponding to the articles, but the English content obtained from web scraping is not the title content.
\\
\hline
\end{tabular}

\label{tab:turenne}
\end{table*}

There are some existing bilingual news datasets in Chinese and English. WikiTitles-v3 \cite{liu-etal-2017-learning} is a dataset of titles. ParaCrawl(bonus) \cite{banon-etal-2020-paracrawl}, WikiMatrix \cite{schwenk-etal-2021-wikimatrix} and BackTrans News \cite{bawden2019university}
  provide parallel corpus in the form of sentences. However, all these databases does not target the financial field.
 By contrast, \cite{Turenne2022mining} provides a Chinese–English parallel dataset which focuses on financial news, using the Financial Times website, from which they grabbed 60,473 news items from between 2007 and 2021. After browsing through the dataset, we discovered that a large number of the Chinese and English texts are not well aligned. Additionally, since the data was scraped from web pages, there are many HTML tags present. We list three examples in Table \ref{tab:turenne}. 

Thus, we aim to create a database exclusively focused on Chinese and English financial news, meticulously proofread by humans to ensure alignment of sentences.

\subsection{Neural Machine Translation}
Neural machine translation (NMT) is a new methodology for machine translation that has led to remarkable improvements. 
Currently there are many existing NMT implementations. Many systems such as those developed in industry by Google, Microsoft, and Baidu, are closed source, and are unlikely to be
released with unrestricted licenses. In addition, we found other open-source neural NMT framework. OpenNMT \cite{klein2018opennmt} is an open-source framework for neural machine translation which can be used to try out new ideas in translation, language modeling, summarization, and many other NLP tasks. So we use OpenNMT to train a model wihch focus on the translation of Chinese and English financial news.

\section{FFN Creation}

We have systematically amassed a substantial volume of financial news articles sourced from various reputable websites, including FOX\footnote{\url{https://www.foxnews.com/}}, CNN\footnote{\url{https://edition.cnn.com/}}, and China Daily\footnote{\url{http://www.chinadaily.com.cn/}}. All these financial news are freely available. The compilation spans a time-frame from January 1, 2014, to December 31, 2023. The dataset can be found in \url{https://github.com/shijing001/FFN_corpus}.

We are committed to crafting a precise and high-quality evaluation dataset. As a result, we refrained from directly scraping sentences from web pages using code. This decision was made because such direct scraping can often result in unaligned text. Therefore, we manually browse web pages, select several paragraphs and the title of a complete news article to add to our dataset, and during the manual screening process, we repeatedly correct the translated results. 

The resulting dataset comprises two distinct categories: main texts, which encompass detailed content within the financial news articles, and titles, representing the headlines of these articles. The identical information is presented in both Chinese (ZH) and English (EN) versions, as delineated in Table \ref{tab:sentence}.
In contrast to the corpora of WMT \cite{kocmi-etal-2022-findings}, our dataset is specifically tailored to financial news, providing content exclusively in simplified Chinese, without the amalgamation of simplified and traditional Chinese characters. Furthermore, when juxtaposed with existing Financial News datasets for text mining \cite{Turenne2022mining}, our dataset, which is manually aligned, ensures translation accuracy and is free of any HTML tags, eliminating the need for further preprocessing. Besides, our dataset stands out for its currency, covering the period from 2014 to 2023, a more recent span compared to the earlier range of 2007 to 2021. Notably, the data in our dataset is sourced from different websites than those in existing datasets, ensuring the provision of distinct data sets even for the same chronological year.

\subsection{Main text}
Main text refers to the primary content within financial news articles, predominantly characterized by lengthy declarative sentences that encompass various clauses. These sentences exhibit a strong contextual meaning. Given the nature of financial news, the inclusion of company names, policy clauses, legal documents, and financial terms is commonplace within these sentences.

It is not sentence-aligned, but paragraph-aligned, which aims to provide the contextual background to examine the influence of context on the translation outcome.

\subsection{Titles}

In contrast to main texts, titles exhibit a distinct nature characterized by brevity and summarization. Essentially, a title serves as a condensed representation or key focal point of the entire article, reflecting a pronounced authorial intent. Notably, titles are often more concise, and some may lack a clear sentence structure, making it inappropriate to categorize them strictly as short sentences. Moreover, the tone employed in titles may lean towards the hyperbolic, strategically designed to captivate readers' attention, thereby differing from the more neutral tone found within paragraph sentences.

It is crucial to note that, as titles are crafted by authors after a comprehensive understanding of the article, their extraction alone may result in an abrupt representation. Additionally, the inherent differences in linguistic thinking between Chinese and English contribute to variations in the titles of the same article across languages.

\section{Experimental Setup}

\begin{table}
\centering
\caption{A list of translation prompts. ZH1 and ZH2 are the results we obtained after translating EN1 and EN2.}
\captionsetup{justification=centering}
\scalebox{1.3}{
\begin{tabular}{cc}
\hline
\multicolumn{2}{c}{Translation Prompt} \\

\hline

  EN1& Translate these sentences\\
& from [SRC] to [TGT]:\\

EN2 & Please provide the [TGT]\\
&translation for these sentences:\\

ZH1& \begin{CJK*}{UTF8}{gbsn}把这些句子从[SRC]翻译成[TGT]：\end{CJK*}\\

ZH2& \begin{CJK*}{UTF8}{gbsn}请为这些句子提供[TGT]的翻译：\end{CJK*}\\
\hline
\end{tabular}

}
\label{tab:prompt}
\end{table}




\subsection{Machine Translation Models}

This comparative study aims to assess the performance of these models in the context of translating Chinese (ZH) to English (EN). By scrutinizing their respective capabilities, we seek to discern any potential advantages or differences in performance, particularly in the realm of ZH-EN translation. This exploration is anticipated to shed light on the strengths and weaknesses of each model, contributing valuable insights to the field of machine translation and language understanding.

For our comparative analysis, we have selected two distinct LLMs: ChatGPT\footnote{\url{https://chat.openai.com/}}, a popular LLM developed by OpenAI, and ERNIE-Bot\footnote{\url{https://yiyan.baidu.com/}} developed by Baidu. Notably, ERNIE-Bot originates from Chinese researchers, prompting our interest in evaluating its efficacy in ZH-EN translation compared to ChatGPT.
For comparison, we also choose DeepL\footnote{\url{https://www.deepl.com/translator}} and Google\footnote{\url{https://translate.google.com/}}, two sets of online translation software.

Additionally, we trained an OpenNMT model \cite{klein2018opennmt} based on the dataset "Financial News dataset for text mining" \cite{Turenne2022mining} and then our dataset serves as its test dataset. We wanted to evaluate this existing dataset, to see how effective it is as a dataset when actually training models.
Because the original author of \cite{Turenne2022mining} did not manually align this dataset, we pre-processed it with manual alignment and removing HTML tags. The resulting database will also be made public and available for research, which can be found in \url{https://github.com/shijing001/FFN_corpus}.

\subsection{Evaluation and Detailed Configuration}

We adopt the BLEU \cite{papineni2002bleu,sundararaman2021syntactic}, TER \cite{snover-etal-2006-study}, chrF \cite{popovic2015chrf} as our evaluation metrics, which is supported by SacreBLEU \cite{post-2018-call}.

In our experiment, we pay attention to the impact of different prompt styles in guiding LLMs’ translation capabilities. We initiated the experiment with two distinct types of English prompts, which were later translated from English to Chinese. As is shown in Table \ref{tab:prompt}. This allowed us to examine whether the prompt's language type affects the translation quality.

\section{Results and Analysis}

\subsection{Performance of various translation systems}

\begin{table*}
\centering
\caption{Performance comparison of five machine translation systems: ChatGPT, ERNIE-Bot, DeepL, Google, and OpenNMT trained from scratch on the dataset  "Financial News dataset for text mining". }
\captionsetup{justification=centering}

\begin{tabular}{c|c|ccc|ccc}
\hline
System & Category & BLEU &TER&chrF& BLEU &TER&chrF  \\
\hline
&&\multicolumn{3}{c|}{ZH-EN}&\multicolumn{3}{c}{EN-ZH}\\
\cdashline{3-8}[2pt/2pt]
\multirow{2}{*}{ChatGPT} & Main text & 22.30 & 70.99&68.64&23.08 &58.14&44.73\\
 & Titles & 15.24&106.08&59.57&17.94&120.75&40.82   \\
\hline
\multirow{2}{*}{ERNIE-Bot} &  Main text & 23.71&69.34&67.53&25.98&55.05&46.26\\
 &Titles & 16.64& 104.40&56.21&24.25&96.30&42.93 \\
\hline
\multirow{2}{*}{DeepL} &Main text &\textbf{29.41}&61.96&\textbf{70.97}&25.89&55.89&45.20\\
&Titles &20.97&74.47&58.10&29.88&60.35&47.92\\
\hline
\multirow{2}{*}{Google} &Main text &27.69&62.72&70.07&\textbf{30.12}&49.91&\textbf{50.08}\\
&Titles &\textbf{34.57}&70.27&\textbf{66.64}&\textbf{34.16}&55.56&\textbf{50.79}\\
\hline
\multirow{2}{*}{OpenNMT model} &Main text &6.53&\textbf{83.61}&42.13&7.75&\textbf{94.14}&25.37\\
&Titles&6.43&\textbf{140.78}&42.82&0.59&\textbf{134.67}&29.70\\
\hline
\end{tabular}

\label{tab:cedg}
\end{table*} 

 Table \ref{tab:cedg} displays the performance of five machine translation systems on both directions (ZH-EN and EN-ZH). Generally, DeepL and Google translation outperform both the ChatGPT and ERNIE-Bot. Especially in the translation of titles, the scores of both translation software are superior to those of the large language model. Particularly in the TER scores for titles, the scores of both translation software (Google Translate and DeepL) clearly demonstrate their superiority in translation accuracy. 
 From this table, the performance of LLMs (ChatGPT and ERNIE-Bot) is quite similar. In terms of translation direction, the performance of LLMs in EN-ZH translation is better than in ZH-EN translation. Overall, the translation quality of the main text is better than that of the titles. 

From Table \ref{tab:cedg}, the BLEU scores of OpenNMT model (trained from scratch) are much lower than those of LLMs and translation software. However, this does not necessarily reflect poor performance of the OpenNMT model itself; rather, it indicates that there are still some issues with the training dataset it relies on. We speculate that the main problem lies in the fact that the dataset itself is too small, and many specialized terms have not been included in it. This actually highlights an issue: there is indeed a shortage of parallel datasets for Chinese and English financial news, and relying solely on the dataset in \cite{Turenne2022mining} is insufficient.

\begin{table*}
\centering
\caption{BLEU scores of LLMs. ZH1, ZH2, EN1, EN2 are those prompts in Table \ref{tab:prompt}. STD represents the standard deviation. AVE means the average score.}
\captionsetup{justification=centering}

\begin{tabular}{c|c|cccc|cc|cccc|cc}
\hline
System & Category & ZH1 & EN1 & ZH2 &EN2&STD&AVE& ZH1 & EN1 & ZH2 &EN2&STD&AVE\\

\hline
&&\multicolumn{4}{c|}{ZH-EN} &&&\multicolumn{4}{c|}{EN-ZH}& \\
\cdashline{1-14}[2pt/2pt]
\multirow{2}{*}{ChatGPT} & Main text& 22.40 & 22.11 & 22.26 & \textbf{22.44}&0.15 &22.30& \textbf{23.40} & 22.62 & 23.12 & 23.18&0.33&23.08\\
 & titles & 14.93 & \textbf{15.86} & 14.99 & 15.16 &0.43&15.24&\textbf{18.31}&17.88&18.13&17.45&0.37&17.94\\
\hline
\multirow{2}{*}{ERNIE-Bot} &  Main text & 23.61 & 23.53 & \textbf{24.03} & 23.68&0.22&23.71&26.12&26.14&\textbf{26.15}&25.53&0.30 &25.98\\
 &titles &\textbf{17.02} & 16.42 & 16.48 & 16.64&0.27&16.64&24.64&\textbf{25.07}&23.10&24.19&0.85&24.25 \\
\hline
\end{tabular}
\label{tab: BLEU ZH-EN EN-ZH LLMs}
\end{table*} 

\begin{table*}
\centering
\caption{Problems of LLMs. "RT" is "The Rejection of Translation", "AMS" is "Answer according to the Meaning of the Sentence", "PY" is "Pinyin Character Feedback", "TC" is "Traditional Chinese Results", "GN" is "Giving Notes", "MO" is "Multiple Outcome", "ROS" is "Reserve the Original Sentences", "IO" is "Information Omission", "EFT" is "Errors in Financial Terminology", "MIS" is "Mispunctuation", "ENCO" is "Errors in the Name of Company and Organization", "TEN" is "Tense", "EM" is "Extended Meaning", "SP" is "Sentence Pattern”.}
\captionsetup{justification=centering}

\begin{tabular}{c|c|cccccccccccccc}
\hline
Category&Model & RT& AMS& PY & TC&GN &MO &ROS & IO & EFT& MIS & ENCO &TEN &EM &SP\\

\hline
\multirow{2}{*}{Main texts}&ChatGPT &&&&&&&& \Checkmark & \Checkmark & \Checkmark & \Checkmark &&& \\
&ERNIE-Bot& \Checkmark& \Checkmark&&&&&& \Checkmark & \Checkmark & \Checkmark & \Checkmark &&& \\
\hline
\multirow{2}{*}{Titles}& ChatGPT & &  & \Checkmark & \Checkmark & \Checkmark &\Checkmark&& & \Checkmark & \Checkmark & \Checkmark & \Checkmark & \Checkmark &\Checkmark\\
&ERNIE-Bot&\Checkmark & \Checkmark &&&\Checkmark&\Checkmark&\Checkmark& & \Checkmark & \Checkmark & \Checkmark & \Checkmark & \Checkmark &\Checkmark\\
\hline
\end{tabular}
\label{tab:unexcpected feedback}
\end{table*} 

\subsection{Performance of \llms over four prompts}

To investigate the effects of prompts on LLMs, we utilize four prompts (two in English and two in Chinese) in Table \ref{tab:prompt}. 
Table \ref{tab: BLEU ZH-EN EN-ZH LLMs} presents the performance of ChatGPT and ERNIE-Bot over those four prompts. Based on the standard deviation of BLUE scores of various prompts, prompts have a certain level of impact on the translation outputs of LLMs. 

\section{Problems of \llms}
To further investigate the specific problems of machine translation of LLMs, we conducted a manual evaluation of the translation results generated by ChatGPT and ERNIE-Bot. Through this evaluation, we discovered the following issues, which is summarized in Table \ref{tab:unexcpected feedback}. 
 For issues unique to ChatGPT, we list them in Table \ref{tab:example.chatgpt}. The detailed explanation for each type of errors are shown as follows. More problematic translation examples of LLMs can be found in the Appendix.

\begin{table*}[ht]
\centering
\caption{Several translation examples from ChatGPT and their error categories.}
\captionsetup{justification=centering}

\begin{tabular}{m{3.8cm}<{\centering}|m{5.2cm}<{\centering}|m{4cm}<{\centering}|m{1.5cm}<{\centering}}
\hline
Original Sentences&Translation Results&Reference&Category\\
\hline
 
Doubt cast on plan to link up with Shanghai stock exchange&\begin{CJK*}{UTF8}{gkai}对计划与上海证券交易所联接产生了质疑。 (Duì jìhuà yǔ Shànghǎi zhèngquàn jiāoyì suǒ liánjiē chǎnshēngle zhìwèn.)\end{CJK*}& \begin{CJK*}{UTF8}{gkai}沪伦通计划引发质疑\end{CJK*}&PY
\\
\hline
Digital economy helps empower China's high-quality development&\begin{CJK*}{UTF8}{gkai}数字经济有助于赋能中国的高质量发展。\end{CJK*} (Simplified Chinese)\begin{CJK*}{UTF8}{bkai}數字經濟有助於賦能中國的高品質發展。
\end{CJK*} (Traditional Chinese)&\begin{CJK*}{UTF8}{gkai}数字经济助力中国高质量发展\end{CJK*} 
&TC\\
\hline
\begin{CJK*}{UTF8}{gkai}迪士尼中国拟发展消费品跨境策略\end{CJK*}&Disney China plans to develop a cross-border consumer products strategy.
\quad \textbf{Note: It is worth mentioning that when translating from Chinese to English, there might be various ways to express the same idea, depending on the desired level of accuracy or context.}&Disney eyes licensed product growth in China&GN\\
\hline
Law to improve financial stability&1. Improving Financial Stability through Legislation
2. Legal measures to enhance financial stability
3. Legislative efforts for improving financial stability
4. Enhancing financial stability through legal means
5. Legislation aimed at improving financial stability
&\begin{CJK*}{UTF8}{gkai}改善金融稳定的法律\end{CJK*}&MO\\
\hline
\end{tabular}
\label{tab:example.chatgpt}
\end{table*}
\textbf{The Rejection of Translation (RT)} \quad On occasions, ERNIE-Bot may decline to translate certain sentences, responding with a message such as "Please refer to relevant websites for more information, and feel free to ask me any other questions." Besides, ERNIE-Bot may provide a translation answer when using one prompt, but rejecting the translation when using another prompt. This indicates that this model is not stable when outputting translation results.

\textbf{Answer according to the Meaning of the Sentence (AMS)} \quad Another observed anomaly in ERNIE-Bot's feedback is its tendency to provide an interpretation or understanding of the given sentences instead of delivering a translation. This behavior is deemed erroneous since the model fails to fulfill the translation request as specified in our prompt. 

\textbf{Pinyin Character Feedback (PY)}\quad In some instances, when prompted in English, ChatGPT may add Pinyin to the results, potentially lowering the overall scores. This could be because ChatGPT assumes that users prompted in English may not understand Chinese, thus including Pinyin to aid pronunciation.

\textbf{Traditional Chinese Results (TC)}\quad Albeit infrequently, when conducting English to Chinese translation with English prompts, ChatGPT may provide results in both simplified and traditional Chinese.

\textbf{Giving Notes (GN)}\quad Sometimes, ChatGPT and ERNIE-Bot may give some notes of the results. This usually does not affect the output of the translation text.

\textbf{Multiple Outcome (MO)}\quad Normally, a single input will result in one translation, but sometimes multiple translations will be given.

\textbf{Reserve the Original Sentences (ROS)}\quad Chances are that ERNIE-Bot may reserve the original sentences rather than translate them.Perhaps due to insufficient training set, ERNIE-Bot cannot translate.

\textbf{Information Omission (IO)} \llms may inadvertently overlook certain information during translation due to an insufficient grasp of contextual nuances. After comprehending the overall meaning of a sentence, the system might erroneously omit certain words, resulting in the loss of crucial information and hindering the reader’s accurate understanding of the original text. This issue is exacerbated when translating long sentences or text with intricate grammatical structures, which strains the system's ability to capture detailed nuances, leading to potential information omission.

\textbf{Errors in Financial Terminology (EFT)} Translation errors in financial terminology are prevalent and significantly impede readers' efficiency and comprehension. These errors often arise from the literal interpretation of technical terms. The underlying cause may be that LLMs lack the corresponding financial terms in their databases, hindering accurate translations. 

\textbf{Mispunctuation (MIS)} \quad The occurrence of such errors primarily stems from the disparity in punctuation conventions between Chinese and English. Chinese employs full-angle punctuation, while English utilizes half-angle punctuation, and many symbols do not have direct equivalents, potentially leading to translation inaccuracies. Furthermore, the divergent grammatical structures of Chinese and English necessitate adjustments during translation, often involving changes in punctuation. If machine translation does not appropriately address these differences, it can result in the incorrect application of punctuation marks, further contributing to translation errors.

\textbf{Errors in the Name of Company and Organization (ENCO)}\quad In the realm of finance, the accurate translation of company names and names of professional organizations holds significant importance. However, \llms often exhibit a tendency to overlook these specific terms, either failing to translate them or providing translations that do not align with the actual names. This oversight can lead to confusion among readers. One plausible explanation for this issue is that language models lack corresponding data in their databases for these specific terms. Additionally, institutions are sometimes presented in the form of abbreviations, and the same abbreviation may have different references in the financial field. In the absence of context, language models may adopt a strategy of not translating to avoid potential inaccuracies in the output.

\textbf{Tense (TEN)} \quad Due to the brevity and contextual limitations inherent in most titles, especially in the context of translation from Chinese to English, \llms may encounter challenges in accurately selecting tenses. This can result in inaccuracies, with past tense phrases being mistakenly rendered as present perfect tense constructions.

\textbf{Extended Meaning (EM)} \quad The textual content of titles often encompasses intricate semantic nuances, integrating elements such as metaphors and personification to convey layers of meaning. However, when processed by \llms for translation, there exists a tendency to prioritize literal interpretations, which can potentially introduce ambiguity into the translated output. This divergence in translation approach may compromise the ability of \llms to accurately capture the nuanced essence of the original title, consequently impacting the clarity and effectiveness of the translated text.

\textbf{Sentence Pattern (SP)} \quad Indeed, a prevalent characteristic of titles is their deviation from complete sentence structures; instead, they commonly feature concise phrases or fragments. However, when subjected to translation by \llms, these titles often undergo an automatic transformation into full sentences, thereby losing their distinctive structural nuances. This transformation can result in a loss of conciseness and impact, ultimately diminishing the effectiveness of the translated title in conveying its intended message. 

Among these problems, Pinyin character feedback, traditional Chinese results, giving notes and multiple outcome can all be avoided by changing the prompts. However, the others actually reflect the translation performance of LLMs themselves, and are not completely eliminated by changing the prompts.

\section{Conclusion}
We have developed a parallel English-Chinese news translation dataset in the finance domain, comprising main texts and titles. Unlike existing datasets, our dataset has been manually verified and revised for high quality, and is current as of December 2023. This dataset can be utilized as a benchmark for evaluating the translation capabilities of LLMs. We observed that various prompts impact LLM translation results, including issues with Pinyin character feedback, traditional Chinese output, annotations, and multiple outcomes. These issues can be mitigated by adjusting the prompts. However, LLMs still exhibit problems such as mispunctuation and errors in company, organization, and financial terminology, highlighting their inherent limitations. Compared to LLMs, translation software like DeepL performs better, especially in translating titles. To enhance LLM competitiveness against translation software, improvements should begin with their training datasets.


\section*{Acknowledgments}

The authors thank the reviewers for the
valuable comments that helped to improve the paper. This work was supported
by the National Natural Science Foundation of
China (grant numbers: 12071302), ``the Fundamental Research Funds for the Central Universities" (grant number 2022114012),  and Mentor Academic Guidance
Program of Shanghai International Studies University (grant number:
2022113028).

\bibliographystyle{IEEEtran}
\bibliography{custom}

\appendix

\begin{table*}[h!]
\centering
\caption{Several examples of the Unexpected feedback of ChatGPT, "Category" represents the type of issue reflected by the translation result, and the abbreviations involved have been detailed in the previous text.}
\captionsetup{justification=centering}

\begin{tabular}{m{3.8cm}<{\centering}|m{5.2cm}<{\centering}|m{4cm}<{\centering}|m{1.5cm}<{\centering}}
\hline
Original Sentences&Translation Results&Reference&Category\\
\hline
 
Doubt cast on plan to link up with Shanghai stock exchange&\begin{CJK*}{UTF8}{gkai}对计划与上海证券交易所联接产生了质疑。 (Duì jìhuà yǔ Shànghǎi zhèngquàn jiāoyì suǒ liánjiē chǎnshēngle zhìwèn.)\end{CJK*}& \begin{CJK*}{UTF8}{gkai}沪伦通计划引发质疑\end{CJK*}&PY
\\
\hline
Digital economy helps empower China's high-quality development&\begin{CJK*}{UTF8}{gkai}数字经济有助于赋能中国的高质量发展。\end{CJK*} (Simplified Chinese)\begin{CJK*}{UTF8}{bkai}數字經濟有助於賦能中國的高品質發展。
\end{CJK*} (Traditional Chinese)&\begin{CJK*}{UTF8}{gkai}数字经济助力中国高质量发展\end{CJK*} 
&TC\\
\hline
\begin{CJK*}{UTF8}{gkai}迪士尼中国拟发展消费品跨境策略\end{CJK*}&Disney China plans to develop a cross-border consumer products strategy.
\quad \textbf{Note: It is worth mentioning that when translating from Chinese to English, there might be various ways to express the same idea, depending on the desired level of accuracy or context.}&Disney eyes licensed product growth in China&GN\\
\hline
Law to improve financial stability&1. Improving Financial Stability through Legislation
2. Legal measures to enhance financial stability
3. Legislative efforts for improving financial stability
4. Enhancing financial stability through legal means
5. Legislation aimed at improving financial stability
&\begin{CJK*}{UTF8}{gkai}改善金融稳定的法律\end{CJK*}&MO\\
\hline
\end{tabular}

\label{tab:example.chatgpt.more}
\end{table*}

\begin{center}
   \begin{table*}[ht]
\centering
\caption{Several examples of the Unexpected feedback of ERNIE-Bot, "Category" represents the type of issue reflected by the translation result, and the abbreviations involved have been detailed in the previous text. }
\captionsetup{justification=centering}

\begin{tabular}{m{3.8cm}<{\centering}|m{5.2cm}<{\centering}|m{4cm}<{\centering}|m{1.5cm}<{\centering}}
\hline
Original Sentences&Translation Results&Reference&Category\\
\hline
 
Doubt cast on plan to link up with Shanghai stock exchange&Please refer to relevant websites for more information, and feel free to ask me any other questions.&\begin{CJK*}{UTF8}{gkai}沪伦通计划引发质疑\end{CJK*}&RT\\
\hline
\begin{CJK*}{UTF8}{gkai}中金股票便宜吗
\end{CJK*} &\begin{CJK*}{UTF8}{gkai}中金股票便宜吗
\end{CJK*} &Is CICC stock cheap?&ROS\\
\hline
\begin{CJK*}{UTF8}{gkai}老龄化、经济放缓、债务累累：中国工业中心地带能否复兴？\end{CJK*}&\begin{CJK*}{UTF8}{gkai}老龄化、经济放缓和债务累累等问题可能会对中国工业中心地带的发展带来一定的影响，但并不意味着这个地区会就此衰落。中国政府已经采取了一系列措施来应对这些问题，包括加强社会保障体系建设、推动创新驱动发展、优化产业结构等。此外，中国拥有庞大的市场和人口规模，这些因素也会为中国经济的发展提供有力的支撑。因此，只要政府和企业积极采取措施，中国工业中心地带仍然有望实现复兴。
\end{CJK*}&Slowing, Graying and in Debt, Can China’s Industrial Heartland Be Revived?&AMS\\
\hline
China and Russia attempt to undercut the US dollar &\begin{CJK*}{UTF8}{gkai}中国和俄罗斯试图削弱美元地位。
这句话的意思是，中国和俄罗斯希望通过采取一系列措施来减少对美国货币的依赖，从而降低美国在全球经济中的影响力。这种努力可能包括推动其他货币的使用、加强本国货币的国际地位、推动贸易和投资多元化等。
\end{CJK*} &\begin{CJK*}{UTF8}{gkai}中国和俄罗斯试图削弱美元\end{CJK*}&GN\\
\hline
\begin{CJK*}{UTF8}{gkai}中俄深化经济联系，抵御西方制裁影响
\end{CJK*}&1. China and Russia are deepening economic ties to better withstand western sanctions pressure.
2. The Chinese and Russian economies are strengthening their ties in order to better resist western sanctions pressure.
3. The Chinese and Russian economies are working together to improve their economic ties to better withstand western sanctions pressure.& China and Russia bind Economies Further, in resisting the impact of Western sanctions&MO\\
\hline
\end{tabular}

\label{tab:example.ernie}
\end{table*} 
\end{center}

\begin{table*}
\centering
\caption{Several examples of the challenges large language models encounter when translating main texts. "Category" represents the type of issue reflected by the translation result, and the abbreviations involved have been detailed in the previous text.}
\captionsetup{justification=centering}
\scalebox{0.81}{
\begin{tabular}{m{4.3cm}<{\centering}|m{4.3cm}<{\centering}|m{4.3cm}<{\centering}|m{1.5cm}<{\centering}|m{3.5cm}<{\centering}}
\hline
Original Sentences&Translation Results&Reference&Category&Notes\\
\hline
 
But a slew of \textbf{hotter-than-expected} economic data reports, including the blowout January jobs report and a disappointing inflation report that pointed to the pervasiveness of high consumer prices, has raised the specter of a higher peak rate or more aggressive rate hikes.&\begin{CJK*}{UTF8}{gkai}但是，一系列经济数据报告显示，包括一份爆炸性的一月份就业报告和一份令人失望的通货膨胀报告，这些报告表明高消费价格的普遍性，引发了更高峰值利率或更积极的加息的担忧。\end{CJK*}&\begin{CJK*}{UTF8}{gkai}但一连串\textbf{热度高于预期的}经济数据报告，包括井喷式的1月就业报告和令人失望的通胀报告，指出了高消费价格的普遍性，使人们对更高的峰值利率或更积极的加息产生了怀疑。\end{CJK*}&IO&In this passage, \textbf{hotter-than-expected} is not translated\\
\hline
That's because of a phenomenon known as "bracket creep," which happens when taxpayers are pushed into higher-income brackets even though their purchasing power is essentially unchanged due to steeper prices for most goods. &\begin{CJK*}{UTF8}{gkai}这是因为一个被称为“阶梯效应”的现象，即纳税人因为大多数商品价格上涨而被推到更高收入税档，即使他们的购买力实际上没有改变。\end{CJK*} &\begin{CJK*}{UTF8}{gkai}这是因为一种被称为“税级攀升”的现象，当纳税人被推入更高税收等级时，由于大多数商品价格上涨，他们的购买力基本没有变化，就会发生这种现象。\end{CJK*}&EFT&\textbf{bracket creep} is a financial term,which should be translated into\begin{CJK*}{UTF8}{gkai}税级攀升\end{CJK*} rather than \begin{CJK*}{UTF8}{gkai}阶梯效应\end{CJK*}\\
\hline
The Inflation Reduction Act passed last year restored tax credits of up to 7500 dollars to buyers of the less expensive Tesla cars, the Model 3 and Model Y, as long as their list price is under 55000 dollars.&\begin{CJK*}{UTF8}{gkai}去年通过的通货膨胀削减法案恢复了高达7,500美元的税收抵免，适用于购买较便宜的特斯拉汽车，包括Model 3和Model Y，只要它们的标价低于55,000美元。\end{CJK*}&\begin{CJK*}{UTF8}{gkai}去年通过的《降低通货膨胀法案》为购买价格较低的特斯拉汽车 Model 3 和 Model Y 的买家恢复了高达 7,500 美元的税收抵免，只要它们的标价低于 55,000 美元。\end{CJK*}&MIS&The Inflation Reduction Act should be translated into\begin{CJK*}{UTF8}{gkai}《通货膨胀减缓法案》\end{CJK*}, which is a document that has a book title mark in Chinese, but \llms did not add it.\\
\hline
As inflation-pinched consumers drink more at home, AB InBev has been shifting the imagery in some of its advertising from traditional settings—such as a bar or sports game—to the home.&\begin{CJK*}{UTF8}{gkai}随着受通货膨胀困扰的消费者在家里多喝酒，AB InBev已经开始在一些广告中调整形象，从传统的场景——比如酒吧或体育比赛——转向家庭。\end{CJK*}&\begin{CJK*}{UTF8}{gkai}由于通货膨胀使消费者在家里喝得更多，百威英博已经将其一些广告中的形象从传统的环境--如酒吧或体育比赛--转移到家里。\end{CJK*}&ENCO&AB InBev's Chinese name is \begin{CJK*}{UTF8}{gkai}百威英博\end{CJK*},but \llms did not translate, remaining the original text intact.\\
\hline
\end{tabular}
}

\label{tab:Several examples of the challenges large language models encounter when translating main texts.}
\end{table*}

\begin{table*}
\centering
\caption{Several examples of the challenges large language models encounter when translating titles. "Category" represents the type of issue reflected by the translation result, and the abbreviations involved have been detailed in the previous text.}
\captionsetup{justification=centering}
\scalebox{0.95}{
\begin{tabular}{m{4cm}<{\centering}|m{4cm}<{\centering}|m{4cm}<{\centering}|m{1.1cm}<{\centering}|m{3.5cm}<{\centering}}
\hline
Original Sentences&Translation Results&Reference&Category&Notes\\

\hline
Year of the tortoise&\begin{CJK*}{UTF8}{gkai}年之龟\end{CJK*}&\begin{CJK*}{UTF8}{gkai}缓慢发展之年\end{CJK*}&EM&
This title employs figurative language in translation, rendering it as "\begin{CJK*}{UTF8}{gkai}缓慢发展之年\end{CJK*}," rather than opting for a literal rendition.\\
\hline
Target sued over LA store stabbings after homeless man attacked woman, 9-year-old boy&\begin{CJK*}{UTF8}{gkai}目标公司因洛杉矶商店刺杀事件被起诉，此前一名无家可归男子袭击了一名妇女和一名9岁男孩。\end{CJK*} &\begin{CJK*}{UTF8}{gkai}塔吉特公司因洛杉矶店铺刺伤案而被起诉，此前一名无家可归男子袭击了一名女性和一名9岁男孩。\end{CJK*} &ENCO&\textbf{Target} is a company,which should be translated into\begin{CJK*}{UTF8}{gkai}塔吉特\end{CJK*} rather than \begin{CJK*}{UTF8}{gkai}目标公司\end{CJK*}\\
\hline
\begin{CJK*}{UTF8}{gkai}巧克力大量短缺\end{CJK*}&There is a severe shortage of chocolate.&The Massive Shortfall of Chocolate&SP&As this is a title, it can be translated into a phrase such as "The Massive Shortfall of Chocolate," rather than being translated as a full sentence.\\
\hline
\begin{CJK*}{UTF8}{gkai}零售业市场竞争加剧\end{CJK*}&The competition in the retail industry market has intensified.&Competition in retail market intensifies&TEN&In the original English text, it employs the simple present tense. Therefore, translating it into the present perfect tense may introduce ambiguity.\\
\hline
Nation's ODI further rises in Jan-Aug&\begin{CJK*}{UTF8}{gkai}国家的日均免打电话增长于1月至8月份。\end{CJK*}&\begin{CJK*}{UTF8}{gkai}1-8月我国对外投资持续增长\end{CJK*}&EFT&"ODI" means \begin{CJK*}{UTF8}{gkai}对外直接投资\end{CJK*} rather than  \begin{CJK*}{UTF8}{gkai}日均免打电话\end{CJK*}.\\
\hline
Hurun Global Rich List released&\begin{CJK*}{UTF8}{gkai}胡润全球富豪榜发布\end{CJK*}&\begin{CJK*}{UTF8}{gkai}《胡润全球富豪榜》发布\end{CJK*}&MIS&Hurun Global Rich List should be translated into\begin{CJK*}{UTF8}{gkai}《胡润全球富豪榜》\end{CJK*}, which is a document that has a book title mark in Chinese, but \llms did not add it.\\
\hline
\end{tabular}
}
\label{tab:Several examples of the challenges large language models encounter when translating titles.}
\end{table*}

\end{document}